# Deep Multiple Kernel Learning

Eric V. Strobl & Shyam Visweswaran

*Department of Biomedical Informatics*
*University of Pittsburgh*
*Pittsburgh, USA*
*evs17@pitt.edu; shv3@pitt.edu*

*Abstract*—Deep learning methods have predominantly been applied to large artificial neural networks. Despite their state-of-the-art performance, these large networks typically do not generalize well to datasets with limited sample sizes. In this paper, we take a different approach by learning multiple layers of kernels. We combine kernels at each layer and then optimize over an estimate of the support vector machine leave-one-out error rather than the dual objective function. Our experiments on a variety of datasets show that each layer successively increases performance with only a few base kernels.

## I. Introduction

Deep learning methods construct new features by transforming the input data through multiple layers of nonlinear processing. This has conventionally been accomplished by training a large artificial neural network with several hidden layers. However, the method has been limited to datasets with very large sample sizes such as the MNIST dataset which contains 60,000 training samples. More recently, there has been a drive to apply deep learning to datasets with more limited sample sizes as typical in many real-world situations.

Kernel methods have been particularly successful on a variety of sample sizes because they can enable a classifier to learn a complex decision boundary with only a few parameters by projecting the data onto a high-dimensional reproducing kernel Hilbert space. As a result, several researchers have investigated whether kernel learning can be modified for deep learning. Cho et al. (2009) described the first approach by optimizing an arc-cosine kernel, a function that mimics the massive random projections of an infinite neural network, and successfully integrated the kernel in a deep architecture. However, the method did not allow easily tunable parameters beyond the first layer. Subsequently, Zhuang et al. (2011) proposed to tune a combination of kernels but had trouble optimizing the network beyond two layers. Moreover, the second layer only consisted of a single Gaussian radial basis function (RBF) kernel.

In this paper, we improve on the previous methods by contributing to several key issues in deep kernel learning. The rest of the paper is structured as follows. First, we describe related work and provide some background on how kernels can be constructed from other kernels. Next, we show that a deep architecture that incorporates multiple kernels can substantially increase the "richness" of representations compared to a shallow architecture. Then, we prove that the upper bound of the generalization error for deep multiple kernels can be significantly less than the upper bound for deep feed-forward networks under some conditions. We then modify the optimization method by tuning over an estimate of the leave-one-out error rather than the dual objective function. We finally show that the proposed method increases test accuracy on datasets with sample sizes as low as the upper tens.

## II. Related Work

Several investigators have tried to extend kernels to deep learning. Cho et al. (2009) described the first approach by developing an arc-cosine kernel that mimics the projections of a randomly initialized neural network. The kernel admits a normalized kernel and can thus be stacked in multiple layers. Successively combining these kernels can lead to increased performance in some datasets. Nonetheless, arc-cosine kernels do not easily admit hyper-parameters beyond the first layer, since the kernel projects the data to an infinite-dimensional reproducing kernel Hilbert space.

Zhuang et al. (2011) attempted to introduce tunable hyper-parameters by borrowing ideas from multiple kernel learning. The authors proposed to successively combine multiple kernels in multiple layers, where each kernel has an associated weight value. However, the authors had trouble optimizing the network beyond a second layer which only consisted of a single Gaussian RBF kernel. In this paper, we improve on the multiple kernel learning approach by successfully optimizing multiple layers each with multiple kernels.

## III. Background

Kernels compute a similarity function between two vector inputs $x, y \in \mathbb{R}^d$. A kernel can be described by the dot product of its two basis functions.

$$K^{(1)}(x,y) = \Phi^{(1)}(x) \cdot \Phi^{(1)}(y),$$

where $K^{(1)}(x,y)$ represents a first layer kernel. One way to view a kernel within a kernel is the respective basis functions within the basis functions for an $l$ number of layers:

$$K^{(l)}(x,y) = \Phi^{(l)}\left(\ldots\Phi^{(1)}(x)\right) \cdot \Phi^{(l)}\left(\ldots\Phi^{(1)}(y)\right).$$

Note that the linear kernel does not change throughout the layers.

$$K^{(l)}(x,y) = \Phi^{(l)}\left(\ldots\Phi^{(1)}(x)\right) \cdot \Phi^{(l)}\left(\ldots\Phi^{(1)}(y)\right)$$
$$= \Phi^{(1)}(x) \cdot \Phi^{(1)}(y) = x \cdot y.$$

In the case of the polynomial kernel, we observe a polynomial of higher order:

$$K^{(1)}(x,y) = (\alpha(x \cdot y) + \beta)^\delta$$

$$K^{(2)}(x,y) = \left(\alpha\left(K^{(1)}(x,y)\right) + \beta\right)^\delta,$$

where $\alpha$, $\beta$ and $\delta$ denote the free parameters of the polynomial kernel. From [1], the Gaussian RBF kernel can be approximated as:

$$K^{(1)}(x,y) \approx K^{(2)}(x,y) = \Phi^{(2)}\left(\Phi^{(1)}(x)\right) \cdot \Phi^{(2)}\left(\Phi^{(1)}(y)\right)$$
$$= e^{-2\gamma(1-K(x \cdot y))}.$$

## IV. COMPLEXITY ANALYSIS

Kernels are designed to create different representations of the data using basis functions. If we stack two kernels of different types, we can often develop a representation that is different from either alone. Moreover, we can obtain "richer" representations that cannot be well-approximated by a single kernel, when we combine multiple kernels within a kernel such as by taking their sum.

More formally, we base an analysis of the richness/complexity of a kernel via its pseudo-dimension and then more specifically by the upper bound of the second-order Rademacher chaos complexity $\hat{U}_n$ as defined in [3]. We also introduce the following new definition:

**Definition 1.** A *deep multiple kernel architecture* is an $l$-level multiple kernel architecture with $h$ sets of $m$ kernels at each layer:

$$K^{(l)} = \left\{\theta^{(l)}_{1,1}K^{(l)}_{1,1}\left(\theta^{(l-1)}_{1,1}K^{(l-1)}_{1,1} + \cdots\right) + \cdots \theta^{(l)}_{h,m}K^{(l)}_{h,m}(\ldots)\right\},$$

where $K^{(l)}_{h,m}$ represents the $m^{\text{th}}$ kernel in set $h$ at layer $l$ with an associated weight parameter $\theta^{(l)}_{h,m}$, and $K^{(l)}$ represents the single combined kernel at layer $l$. The term $\mathcal{K}^{(l)}$ is used as short-hand to denote all kernels in layer $l$. The architecture is depicted in Figure 1.

**Theorem 1.** Let $m$ be a finite number of base kernels, $\mathcal{K}^{(1)}$ the single layer kernel functions, and $\mathcal{K}^{(l>1)}$ the multi-layer kernel functions. Then:

$$\hat{U}_n(\mathcal{K}^{(1)}) \leq \hat{U}_n(\mathcal{K}^{(l>1)}).$$

**Proof.** The tunable weights of the first and last layer can be represented as two second-order tensors of non-negative $\mathbb{R}^{h \times m}$. Assuming the same architecture for each layer excluding the first and last, the number of weights can be represented as a fourth-order tensor of non-negative $\mathbb{R}^{(l-2) \times h \times h \times m}$. The total number of free weights in $\mathcal{K}^{(l>1)}$ is thus $(l-2)h^2m + 2hm$. The pseudo-dimension of $\mathcal{K}^{(l>1)}$ can now be stated as $d_\mathcal{K} \leq (l-2)h^2m + 2hm$. On the other hand, the pseudo-dimension of $\mathcal{K}^{(1)}$ for the single layer kernels can be stated as $d_\mathcal{K} \leq m$ (Lemma 7, [4]). We can now derive the upper bound of the Rademacher chaos complexity for the single and multi-layer cases from Theorem 3, [3]:

$$\hat{U}_n(\mathcal{K}^{(1)}) \leq (192e + 1)u^2m$$
$$\hat{U}_n(\mathcal{K}^{(l>1)}) \leq (192e + 1)u^2((l-2)h^2m + 2hm),$$

where $e$ is a natural constant, and $u := \sup_{K \in \mathcal{K}, x \in X} \sqrt{K(x,x)}$. Thus, $\hat{U}_n(\mathcal{K}^{(1)}) \leq \hat{U}_n(\mathcal{K}^{(l>1)})$. □

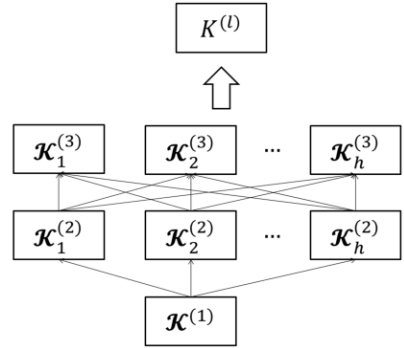

Fig 1. Depiction of a deep multiple kernel architecture. Lines represent the weights for each set, $\theta^{(l)}_h$.

**Remark.** The looser upper bound with a deep compared to a shallow multiple kernel architecture suggests that multiple layers can increase the richness of the kernel representations.

## V. COMPARISON TO FEED-FORWARD NETWORKS

The increased richness of the kernels can increase the risk of over-fitting. However, we can prove that the upper bound of the generalization error for deep multiple kernels is significantly less than the upper bound for deep feedforward networks under some conditions.

**Definition 2.** We define a *large margin feed-forward network* in which a large margin classifier is applied to the last hidden layer of the network. We can thus equivalently represent this feed-forward network in kernel form. We define the large margin feed-forward network for an instance as $F(x_i)$ and its kernel as:

$$K(F(x_i), F(x'_i)) = F(x_i) \cdot F(x'_i).$$

**Theorem 2.** The $\hat{U}_n$ upper bound of the deep multiple kernel is proportional to $u$ with the $\hat{U}_n$ upper bound of the large margin feed-forward network kernel when:

$$d = \sqrt{\frac{(l-2)h^2m + 2hm}{(l-1)}},$$

where $d$ represents the dimensionality of the data and the number of hidden nodes at each layer.

**Proof.** Assuming we adopt the same number of hidden nodes as the dimensionality of the data, the weights of the large margin feed-forward network can be represented as a third-order tensor, where the number of free parameters is $d^2(l-1)$. We equate the number of free parameters from the feed-forward network kernel to the number of free parameters of a deep multiple kernel as derived in Theorem 1 assuming the same number of layers.

$$d^2(l-1) = (l-2)h^2m + 2hm,$$

$$d = \sqrt{\frac{(l-2)h^2m + 2hm}{(l-1)}}.$$

In this case, both the large margin feed-forward network kernel and the deep multiple kernel have the same pseudo-dimension upper bound. Hence, it follows that both have a Rademacher chaos complexity upper bound proportional to $u$ from Theorem 1, [3]. □

**Remark.** Theorem 2 implies that a deep multiple kernel can have a lower generalization bound than a large margin feed-forward network kernel, if we select a small number of base kernels and sets of base kernels at each layer. This is in contrast to the large feed-forward networks traditionally used in deep learning.

## VI. OPTIMIZATION METHOD

The classifier given by an SVM is $sgn(\sum_{i=1}^{l} \alpha_i^0 y_i K_\theta(x_i,x) + b)$. Ideally, we would like to choose the coefficients $\alpha^0$ to minimize an estimate of the true risk of the SVM. Traditionally, this has been solved by maximizing the margin through the gradient of the dual objective function with respect to the kernel hyper-parameters. However, deep learning schemes present a risk of over-fitting with increased richness of the representations. Thus, it is particularly important to seek a tight bound of the leave-one-out error. In this paper, we decided to use the span bound, since it has shown promising results in single layer multiple kernel learning [5]. Assuming that the set of support vectors remains the same throughout the leave-one-out procedure, the span bound can be stated as:

$$L((x,y),\dots,(x,y)) \leq \sum_{p=1}^{l} \phi(\alpha_p^0 S_p^2 - 1) =: T_{Span},$$

where $L$ is the leave-one-out error, and $S_p$ is the distance between the point $\Phi_{K_\theta}(x_p)$ and the set $\Gamma_p = \{\sum_{i\neq p, \alpha_i^0 > 0} \lambda_i \Phi_{K_\theta}(x_i) \mid \sum_{i\neq p} \lambda_i = 1\}$.

We now modify the arguments presented in Liu et al. (2011) for deep multiple kernel learning. The estimate of the span bound requires a step function that is not differentiable. Therefore, we can smooth the step function instead by using a contracting function $\phi(x) = (1 + \exp(-cx + b))^{-1}$, where $c$ and $d$ are non-negative weights. Similar to [6], we chose $c = 5$ and $d = 0$. Chapelle et al. (2002) showed that $\bar{S}_p^2$ can then be smoothed by adding a regularization term:

$$\bar{S}_p^2 = \min_{\lambda, \sum \lambda_i = 1} \left\| \Phi_{K_\theta}(x_p) - \eta \sum_{i\neq p}^{l} \lambda_i \Phi_{K_\theta}(x_i) \right\| + \eta \sum_{i\neq p}^{l} \frac{1}{\alpha_i^0} \lambda_i^2.$$

Now, denote the set of support vectors sv $= \{x | \alpha_i^0 > 0, i = 1,\dots,l\}$, $\widetilde{\mathcal{K}}_{\theta_{sv}} = \begin{pmatrix} \mathcal{K}_{\theta_{sv}} & \mathbf{1} \\ \mathbf{1}^T & 0 \end{pmatrix}$ and $\frac{\partial \widetilde{\mathcal{K}}_{k_{sv}}^0}{\partial \theta_k} = \begin{pmatrix} \frac{\partial \widetilde{\mathcal{K}}_{k_{sv}}}{\partial \theta_k} & \mathbf{0} \\ \mathbf{0}^T & 0 \end{pmatrix}$. With these new notations, we can rewrite $\bar{S}_p^2$ as $1/(\widetilde{\mathcal{K}}_{\theta_{sv}} + Q)_{pp}^{-1} - Q_{pp}$, where $Q$ is a diagonal matrix with elements $[Q]_{ii} = -\eta/\alpha_i^0$ and $[Q]_{n_{sv}+1,n_{sv}+1} = 0$.

**Theorem 3.** Let $G$ be a diagonal matrix with elements $[G]_{ii} = -\eta/(\alpha_{sv_i}^0)^2$ and $[G]_{n_{sv}+1,n_{sv}+1} = 0$. We also define $\bar{A}$ as the inverse of $\widetilde{\mathcal{K}}_{\theta_{sv}}$ with the last row and column removed. Then,

$$\frac{\partial \bar{S}_p^2}{\partial \theta_k} = \frac{1}{B_{pp}^{-1}} \left( B^{-1} \left( \frac{\partial \widetilde{\mathcal{K}}_{k_{sv}}^0}{\partial \theta_k} + GF \right) B^{-1} \right)_{pp} - (GF)_{pp},$$

where $B = \widetilde{\mathcal{K}}_{\theta_{sv}} + Q$, $Y_{sv} = diag\left((y_{sv_1},\dots,y_{sv_n})^T\right)$ and $F = diag(Y_{sv}\bar{A}\frac{\partial \mathcal{K}_{k_{sv}}}{\partial \theta_k}Y_{sv}\alpha_{sv}^0; 0)$. The proof can be found in [5]. □

We calculate $\frac{\partial \widetilde{\mathcal{K}}_{\theta_{sv}}}{\partial \theta_k}$ by performing the standard chain rule, where each set is normalized to a unit hypersphere:

$$\mathcal{K}_h^{(l)}(x,y) \leftarrow \frac{\mathcal{K}_h^{(l)}(x,y)}{\sqrt{\mathcal{K}_h^{(l)}(x,x)\mathcal{K}_h^{(l)}(y,y)}}.$$

Normalization is critical to prevent kernel values from growing out of control. We can now create an algorithm with the derivative of $\partial T_{Span}/\partial \theta$ by alternating between (1) fixing $\alpha$ and solving for $\theta$, and (2) fixing $\theta$ and solving for $\alpha$.

| **Algorithm**: Adaptive Span Deep Multiple Kernel Learning Algorithm |
|---|
| 1. **Input:** $\theta_k^1$ and $\gamma_k \in [0,1]$ for every kernel $k$ |
| 2. **for** $t = 1,2,\dots$ **do** |
| 3.     solve the SVM problem with $K^{(l)}(\theta^t)$ |
| 4.     **for** $k = 1,2,\dots$ **do** |
| 5.         $\theta_k^{t+1} \leftarrow \theta_k^t - \gamma_k \frac{\partial T_{Span}}{\partial \theta_k}$ |
| 6.     **end for** |
| 7.     **if** stopping criterion **then** break |
| 8. **end for** |

## VII. EXPERIMENTS

Multiple kernel learning algorithms have traditionally used RBF and polynomial kernels. However, we chose not to use these kernels, since our objective based on the proposed theorems suggests that we should try to maximize the upper bound of the pseudo-dimension of the final kernel to increase its richness with each successive layer. In fact, it can be shown that the sum of RBF kernels has a pseudo-dimension of 1 from Lemma 2, [3]. Hence, we use four unique base kernels: a linear kernel, an RBF kernel with $\gamma = 1$, a sigmoid kernel with $\alpha = -1 \times 10^{-4}$ and $\beta = 1$, and a polynomial kernel with $\alpha = 1$, $\beta = 1$, $\delta = 2$. We used one set of kernels for each layer making the 3-layer Radamacher upper bound of the architecture proportional to a large margin feed-forward network kernel with $d = \sqrt{6}$ according to Theorem 2. We initialize all $\theta_1^{(l)}$ to $\frac{1}{m}$. Moreover, we use gradient descent on the span bound for 500 iterations for both shallow and deep multiple kernel architectures with $C$ fixed to 10 on 22 standardized UCI datasets. Datasets were randomized and split in half, while instances with missing values were excluded.

| Dataset | n/d | n | Dual ||||  Span |||
|---|---|---|---|---|---|---|---|---|---|
|  |  |  | *1-Layer* | *Zhuang* | *2-Layer* | *3-Layer* | *1-Layer* | *2-Layer* | *3-Layer* |
| Arcene | 0.01 | 100 | 83.00 | 80.00 | 83.00 | 83.00 | **84.00** | 83.00 | 83.00 |
| Musk1 | 1.43 | 238 | 94.12 | 94.96 | 94.96 | 95.38 | 94.96 | **95.80** | **95.80** |
| Sonar | 1.73 | 104 | 89.42 | 88.46 | 89.42 | 89.42 | 88.46 | **90.38** | 89.42 |
| Indian Liver | 2.90 | 290 | 65.52 | 68.97 | 66.55 | 67.24 | 68.38 | 70.34 | **70.69** |
| Zoo | 3.19 | 51 | 92.16 | 92.16 | 92.16 | 92.16 | **94.12** | 92.16 | 92.16 |
| Ionosphere | 5.18 | 176 | 90.91 | 91.48 | 93.75 | 94.32 | 90.91 | 92.61 | **94.89** |
| Post-Operative | 5.38 | 43 | 55.81 | **65.12** | 55.81 | 60.47 | 55.81 | 55.81 | 55.81 |
| Audiology | 7.71 | 54 | **56.60** | 54.72 | 54.72 | 50.94 | 52.83 | 54.72 | 54.72 |
| Glass2 | 9.00 | 81 | 69.14 | 70.37 | 67.90 | 70.37 | 71.60 | **75.31** | **75.31** |
| Corral | 10.67 | 64 | **100.0** | **100.0** | **100.0** | **100.0** | **100.0** | **100.0** | **100.0** |
| Cleve | 12.17 | 146 | 73.29 | 73.97 | **74.66** | **74.66** | 70.55 | 73.29 | 73.29 |
| Congress | 13.63 | 218 | **94.95** | 94.04 | **94.95** | **94.95** | 94.04 | **94.95** | **94.95** |
| Credit | 21.80 | 327 | 81.96 | 83.49 | 82.26 | **84.40** | **84.40** | **84.40** | **84.40** |
| Australian | 24.64 | 345 | 80.29 | 81.45 | 82.03 | 81.45 | **82.32** | **82.32** | **82.32** |
| German | 25.00 | 500 | 69.60 | 70.80 | **71.20** | 69.40 | 68.40 | 69.60 | 69.40 |
| 3of9 | 28.44 | 256 | **99.61** | 98.83 | 99.22 | 99.22 | 98.83 | 99.22 | 99.22 |
| Liver | 28.67 | 173 | 67.05 | 68.21 | 70.52 | **71.68** | 70.52 | 71.10 | 70.52 |
| Monk3 | 36.00 | 216 | 64.35 | 64.81 | 64.81 | **69.44** | 68.98 | **69.44** | **69.44** |
| Breast Cancer | 38.11 | 343 | 97.67 | **98.54** | 97.96 | **98.54** | 97.67 | 97.96 | 97.96 |
| Pima Indians | 48.00 | 384 | 70.57 | 76.56 | 77.10 | 76.82 | **78.65** | 77.10 | 77.60 |
| Tic-Tac-Toe | 53.22 | 479 | **95.40** | 92.07 | 92.90 | 91.44 | 87.89 | 92.90 | 92.90 |
| Balance Scale | 72.00 | 288 | 98.61 | 98.26 | 98.61 | 98.61 | **99.65** | 98.96 | 98.96 |
| Rank |  |  | 3.18 | 2.73 | 2.50 | 2.32 | 2.64 | 1.91 | 1.82 |
| p-value |  |  | **0.022** | **0.018** | 0.083 | 0.340 | **0.047** | 1.000 |  |

Table 1. Percent accuracies after optimizing over the dual objective function or span bound. n/d stands for training sample size over dimensions; Zhuang for Zhuang et al. (2011). The p-values were obtained by comparing against span 3-layer by paired Wilcoxon signed-rank test.

We show increased accuracy with the incorporation of each successive layer by optimizing over the dual objective and span bound. There was a larger increase in accuracy with the addition of the second layer than with the addition of the third. However, the third layer did result in small increases in accuracy such as the 2% increase seen in the Ionosphere dataset with the span bound.

The proposed method increases accuracy on a range of sample sizes. The experimental results are thus consistent with the theorems proposed in section V. Namely, we can avoid over-fitting by choosing a small number of base kernels and sets of kernels at each layer. Thus, similar to single layer kernels, the key to increased accuracy may be to choose a few appropriate kernel representations. At the very least, we can choose a set of appropriate single layer kernels and then the deep architecture can help boost accuracy beyond the single layer.

The method of optimizing over the span bound generally performs better than optimizing over the dual objective function. The performance difference is significant as the 2-layer optimized over the span outperforms the 3-layer optimized over the dual. These results are consistent with the conclusions in Section VI that using a tighter upper bound on the generalization performance can help offset the increased kernel complexity with each subsequent layer.

## VIII. CONCLUSION

We have developed a new method to successfully optimize multiple, complete layers of kernels while increasing generalization performance on a variety of datasets. The method works by combining multiple kernels within each layer to increase the richness of representations and then by optimizing over a tight upper bound of the leave-one-out error.

ACKNOWLEDGEMENTS

This research was funded by the NLM/NIH grant T15 LM007059-24 to the University of Pittsburgh Biomedical Informatics Training Program and the NIGMS/NIH grant T32 GM008208 to the University of Pittsburgh Medical Scientist Training Program.